\documentclass{ecai2004}
\usepackage{times}
\usepackage{graphicx}
\usepackage{latexsym}

\def\alcd{{\cal ALC}({\cal D})}
\def\rcc8{\mbox{${\cal RCC}$8}}
\def\alc{{\cal ALC}}
\def\st{\mbox{s-t}}
\def\xml{\mbox{\em XML}}
\def\alcf{{\cal ALCF}}
\def\xdl{{\cal MTALC}({\cal D}_x)}
\def\xdlplus{{\cal MTALC}({\cal D}_x,{\cal D})}
\def\wrt{\mbox{w.r.t.}}
\def\atra{{\cal CYC}_t}
\def\rccdc{{\mbox{\em DC}}}
\def\rccec{{\mbox{\em EC}}}
\def\rcctpp{{\mbox{\em TPP}}}
\def\rccpo{{\mbox{\em PO}}}
\def\rcceq{{\mbox{\em EQ}}}
\def\rccntpp{{\mbox{\em NTPP}}}
\def\rcctppi{{\mbox{\em TPPi}}}
\def\rccntppi{{\mbox{\em NTPPi}}}
\def\deuxdo{\mbox{2D}{\cal O}}
\def\apra{{\cal CYC}_b}
\def\rtopspace{{\cal RTS}}
\def\rccats{\mbox{$\rcc8${\em -at}}}
\def\atraats{\mbox{$\atra${\em -at}}}
\def\topspace{{\cal TS}}
\def\xat{{\mbox{x-{\em at}}}}
\def\mtalc{{\cal MTALC}}
\def\pltl{{\cal PLTL}}
\def\xdlpluspq{{\cal MTALC}_{p,q}({\cal D}_x,{\cal D})}
\def\mtalcpq{{\cal MTALC}_{p,q}}
\def\twcatac{\mbox{\em twc-atac}}
\def\dnf{\mbox{\em dnf}}
\def\csp{\mbox{\em CSP}}
\def\queue{\mbox{\em Queue}}
\begin{document}
\title{\underline{\footnotesize{in Proceedings of the ECAI Workshop on Spatial and
Temporal Reasoning, pp. 123-127, Valencia, Spain, 2004:}}\\
Augmenting $\alcd$ (atemporal) roles and (aspatial) concrete 
domain with temporal roles and a spatial concrete domain -first
results}

\author{Amar Isli\\
Fachbereich Informatik, Universit\"at Hamburg\\
am99i@yahoo.com}

\maketitle
\bibliographystyle{ecai2004}
\underline{WORK EXACTLY AS REJECTED AT THE MAIN ECAI}\footnote{European
Conference on Artificial Intelligence.}\underline{ 2004}\footnote{The
\underline{reviews} are added to the actual paper, after the references,
for potential people interested in objectivity of conferences' reviewing
processes.}
\begin{abstract}
Consider an $\alcd$ (tree-like) interpretation ${\cal I}$: a node of ${\cal I}$ can
be seen as labelled with a set of atomic concepts (atomic
propositions), and pairs of the form $(g,O)$ where $g$ is a function
representing a concrete feature and $O$ a value from the universe of
instantiation values of the concrete domain. ${\cal I}$ describes thus
(structured) conceptual knowledge, on the one hand, and, on the
other hand, an instantiation of the variables (concrete features)
with concrete values of the concrete domain. $\alcd$ does not assume
restrictions (specialisations) of the roles, nor of the concrete
domain: the roles are considered atemporal, and the concrete
domain aspatial. ${\cal I}$ can be seen as a snapshot of the World at a
specific moment of time, i.e., as a situation in the situation
calculus terminology. Consider now an interpretation $J$ of a modal
temporal logic, assigning at each time point a truth value with each
element of a set $P$ of atomic propositions. The atomic propositions
can be seen as atomic conceptual knowledge. To make such
interpretations $J$ richer, with each time point is associated an
$\alcd$ interpretation as described above, instead of just atomic
propositions. We can go even further, by considering a (dynamic)
spatial scene with, say, $n$ objects: we can then make the $J$
interpretation even richer: with each time point is associated, not
only an $\alcd$ interpretation, describing the look of the conceptual
knowledge at that point (conceptual situation), but the description
of the spatial scene at that point (spatial situation) as well,
either by giving the positions of the different objects of the
scene, or the spatial relations on tuples of objects of the scene,
such as, e.g., $\rcc8$ relations on pairs of the objects, if the scene
consists of regions of a topological space. We provide first results
on a framework handling such rich structures, and obtained by
augmenting $\alcd$ atemporal roles and aspatial concrete domain with
temporal roles and a spatial concrete domain.

Keywords: Description Logics, Temporal Reasoning, Spatial Reasoning, 
Reasoning about Actions and Change, Constraint Satisfaction, Knowledge 
Representation, Qualitative Reasoning, Situation Calculus.
\end{abstract}
\newtheorem{definition}{Definition}
\section{Introduction}\label{sectone}
The well-known $\alcd$ family of description logics (DLs) with a
concrete domain \cite{Baader91a} originated from a pure DL known
as $\alc$ \cite{Schmidt-Schauss91a}, with $m\geq 0$ roles all of which are
general, not necessarily functional relations. It is obtained by
adding to $\alc$ functional roles (better known as abstract features), a
concrete domain ${\cal D}$, and concrete features (which refer to objects of
the concrete domain).

Consider now the family of domain-specific spatio-temporal (henceforth $\st$)
languages, obtained by spatio-temporalising $\alcd$ in the following way:
\begin{enumerate}
  \item temporalisation of the roles, so that they consist of $m+n$
    immediate-successor (accessibility) relations
    $R_1,\ldots R_m,f _1,\ldots ,f _n$, of which the $R_i$'s are general, not
    necessarily functional relations, and the $f_i$'s functional relations;
    and
  \item spatialisation of the concrete domain ${\cal D}$: the spatialisation
    is ${\cal D}_x$, generated  by a spatial Relation Algebra (RA) $x$, such as the
    Region-Connection Calculus RCC8 \cite{Randell92a}.
\end{enumerate}
The resulting family, together with what we will refer to as weakly cyclic
TBoxes, enhances the expressiveness of modal temporal logics with qualitative spatial
constraints, and consists of qualitative theories for (relational) spatial change
and propositional change in general, and for motion of spatial scenes in particular.
In particular, satisfiability of a concept with respect to ($wrt$) a weakly cyclic TBox is decidable.

An interpretation of a member of such a spatio-temporal family is a (labelled)
tree-like structure. A snapshot of such a structure (i.e., the label of a node)
describes a static situation, splitting into a propositional (sub-)situation,
given by the set of atomic propositions true at that node, and a (relational)
spatial (sub-)situation, given by a consistent conjunction of qualitative
spatial relations on tuples of concrete features (the qualitative spatial relations
are predicates of the concrete domain).\footnote{We could use an instantiation of
concrete values to the concrete features, but knowing that such an instantiation
exists, given consistency of the conjunction of qualitative constraints, is enough.}
Real applications, however, such as high-level vision, $\xml$ documents, or what is
known as spatial aggregation (see, e.g., \cite{BaileyKellogg03a}), have a huge demand in
the representation of dynamic structured data. Such structured data may consist of
descriptions of complex objects, or of classes of objects, such as, e.g., a complex
table setting for a meal, a tree-like description of a complex $\xml$ document, or a
complex spatial aggregate.

We denote by $\alcf$ the DL $\alc$ \cite{Schmidt-Schauss91a}
augmented with abstract features. $\alcf$ is particularly
important for the representation of static structured data, thanks, among
other things, to its abstract features, which allow it to access
specific paths. $\alcf$ is a sublanguage of $\alcd$, making the latter also
suitable for the representation of static structured data. $\alcf$,
however, contrary to $\alcd$, does not allow for the representation of
domain-specific knowledge, which can be seen as constraints on objects of
the domain of interest, and which $\alcd$ is very good at, thanks to its
concrete domain.

The roles in $\alcd$ are interpreted in the same way as inheritance
relations in semantic networks; in particular, they are given no tenporal
interpretation. The concrete domain is just an abstract constraint language;
in particular the universe of instantiation values is given no spatial
interpretation: if a constraint on $X$ and $Y$ is seen as a binary Boolean
matrix then value $1$ in entry $(i,j)$ means that assigning the $i$-th value
of the universe to $X$ matches with assigning the $j$-th value to $Y$. In
other words, the constraint does not say anything about how the arguments
relate, say, spatially to each other, which would be different if the
relations were, say, $\rcc8$ relations (and the universe of instantiation
values, regions of a topological space). As such, $\alcd$ describes
structured static data, with the possibility of expressing domain specific
constraints, thanks to its concrete domain.

We denote the $\alcd$ spatio-temporalisation referred to above as $\xdl$
(Modal Temporal $\alc$ with a concrete domain generated by spatial RA $x$).
The roles are now given a temporal interpretation, and they consist of
immediate-successor relations (functional relations in the case of abstract
features, and general relations in the case of non-functional roles); they
can be seen as actions in the possible-worlds semantics of the situation
calculus (see, e.g., \cite{Sandewall96b}).

The extension of $\alcd$ we will be considering in this work is indeed a
cross product of the spatio-temporalisation $\xdl$, on the one hand, and
$\alcd$ itself, on the other hand. It will be referred to as $\xdlplus$.
Section \ref{secttwo} provides a brief background on the spatial relations to be
used as predicates of the spatial concrete domain. Section \ref{sectthree} briefly
describes an aspatial concrete domain, as the $\alcd$ one. Section \ref{sectfour}
describes the spatial concrete domains to be used in the paper. The syntax
of $\xdlplus$ concepts is given in Section \ref{sectfive}. Weakly cyclic TBoxes and
the $\xdlplus$ semantics will be described in Sections \ref{sectsix} and \ref{secteight},
respectively. An overview of decidability of the problem of satisfiability
of an $\xdlplus$ concept $\wrt$ a weakly cyclic TBox will be given in
Section \ref{sectnine}.

We first provide some background on binary relations.
Given a set $A$, we denote by $|A|$ the cardinality of $A$.
A binary relation, $R$, on a set $S$ is any subset of the cross product
$S\times S=\{(x,y):x,y\in S\}$.
Such a relation is reflexive $\iff$ $R(x,x)$, for all $x\in S$;
it is symmetric $\iff$, for all $x,y\in S$, $R(y,x)$, whenever $R(x,y)$;
it is transitive $\iff$, for all $x,y,z\in S$, $R(x,z)$, whenever $R(x,y)$ and $R(y,z)$;
it is irreflexive $\iff$, for all $x\in S$, $\neg R(x,x)$;
it is antisymmetric $\iff$, for all $x,y\in S$, if $R(x,y)$ and $R(y,x)$ then $y=x$; and
it is serial $\iff$, for all $x\in S$, there exists $y\in S$ such that $R(x,y)$.
The transitive (resp. reflexive-transitive) closure of $R$ is the smallest relation
$R^+$ (resp. $R^*$), which includes $R$ and is transitive (resp. reflexive and
transitive).
Finally, $R$ is functional if, for all $x\in S$, $|\{y\in S:R(x,y)\}|\leq 1$;
it is nonfunctional otherwise.
\section{A brief background on $\rcc8$ and $\atra$}\label{secttwo}
{\bf The RA $\rcc8$.}
The RCC-8 calculus \cite{Randell92a} consists of a set of eight
JEPD (Jointly Exhaustive and Pairwise Disjoint) atoms, $\rccdc$ (DisConnected), $\rccec$ (Externally Connected),
$\rcctpp$ (Tangential Proper Part), $\rccpo$ (Partial Overlap), $\rcceq$
(EQual),
$\rccntpp$ (Non Tangential Proper Part), and the converses, $\rcctppi$ and
$\rccntppi$, of $\rcctpp$ and $\rccntpp$, respectively.

\noindent {\bf The RA $\atra$.}
The set $\deuxdo$ of 2D orientations is defined in the usual way, and is isomorphic to the set of directed
lines incident with a fixed point, say $O$. Let $h$ be the natural isomorphism, associating with each
orientation $x$ the directed line (incident with $O$) of orientation $x$. The angle $\langle x,y\rangle$
between two orientations $x$ and $y$ is the anticlockwise angle $\langle h(x),h(y)\rangle$.
The binary RA of 2D orientations in \cite{Isli00b}, $\apra$, contains four atoms:
$e$ (equal), $l$ (left), $o$ (opposite) and $r$ (right). For all $x,y\in\deuxdo$:
$e(y,x) \Leftrightarrow \langle x,y\rangle =0$;
$l(y,x) \Leftrightarrow \langle x,y\rangle\in (0,\pi )$;
$o(y,x) \Leftrightarrow \langle x,y\rangle =\pi$;
$r(y,x) \Leftrightarrow \langle x,y\rangle\in (\pi ,2\pi )$.
Based on $\apra$, a ternary RA, $\atra$, for cyclic ordering of 2D
orientations has been defined in \cite{Isli00b}: $\atra$ has $24$ atoms, thus $2^{24}$
relations.
The atoms of $\atra$ are written as $b_1b_2b_3$, where
$b_1,b_2,b_3$ are atoms of $\apra$, and such an atom
is interpreted as follows:
$(\forall x,y,z\in\deuxdo )(b_1b_2b_3(x,y,z)\Leftrightarrow b_1(y,x)
 \wedge b_2(z,y)\wedge b_3(z,x))$.
The reader is referred to \cite{Isli00b} for more details.
\section{The $\alcd$ aspatial concrete domain}\label{sectthree}
The role of a concrete domain in so-called DLs with a concrete domain \cite{Baader91a}, is to give
the user of the DL the opportunity to represent, thanks to predicates, knowledge on
objects of the application domain, as constraints on tuples of these objects.
\begin{definition}[concrete domain \cite{Baader91a}]\label{cddefinition}
A concrete domain ${\cal D}$ consists of a pair
$(\Delta _{{\cal D}},\Phi _{{\cal D}})$, where
$\Delta _{{\cal D}}$ is a set of (concrete) objects, and
$\Phi _{{\cal D}}$ is a set of predicates over the objects in $\Delta _{{\cal D}}$.
Each predicate $P\in\Phi _{{\cal D}}$ is associated
with an arity $n$: $P\subseteq (\Delta _{{\cal D}})^n$.
\end{definition}
\begin{definition}[admissibility \cite{Baader91a}]\label{cdadmissibility}
A concrete domain ${\cal D}$ is admissible if:
(1) the set of its predicates is closed under negation and
    contains a predicate for $\Delta _{{\cal D}}$; and
(2) the satisfiability problem for finite conjunctions of
    predicates is decidable.
\end{definition}
\section{The spatial concrete domains ${\cal D}_x$,
            with $x\in\{\rcc8 ,\atra\}$}\label{sectfour}
The concrete domain generated by $x$, ${\cal D}_x$, can be written as
${\cal D}_x=(\Delta _{{\cal D}_x},\Phi _{{\cal D}_x})$, with
$                         {\cal D}_{\rcc8}   =(\rtopspace ,2^{\rccats})$ and
$                         {\cal D}_{\atra}   =(\deuxdo ,2^{\atraats})$,
where:
\begin{enumerate}
  \item $\rtopspace$ is the set of regions of a topological space $\topspace$;
    $\deuxdo$ is the set of 2D orientations; and
  \item $\xat$ is the set of $x$ atoms
    ---$2^{\xat}$ is thus the set of all $x$ relations.
\end{enumerate}
Admissibility of the concrete domains ${\cal D}_x$ is a direct consequence of
(decidability and) tractability of the subset $\{\{r\}|r\in\xat\}$ of $x$
atomic relations (see \cite{Renz99b} for
$x=\rcc8$, and \cite{Isli00b} for $x=\atra$).
\section{Syntax of $\xdlplus$ concepts,
            with $x\in\{\rcc8 ,\atra\}$}\label{sectfive}
\begin{definition}\label{defxdlconcepts}
Let $x$ be an RA from the set $\{\rcc8 ,\atra\}$. Let $N_C^{at}$,
$N_C^t$, $N_R^{at}$, $N_R^t$, $N_{cF}^{as}$ and $N_{cF}^s$ be mutually
disjoint and countably infinite sets of atemporal concept names, temporal
concept names, atemporal role names, temporal role names, aspatial
concrete features, and spatial concrete features, respectively;
$N_{aF}^{at}$ a countably infinite subset of $N_R^{at}$ whose elements are
atemporal abstract features; and
$N_{aF}^t$ a countably infinite subset of $N_R^t$ whose elements are
temporal abstract features. A spatial (concrete)
feature chain is any finite composition $f_1^t\ldots f_n^tg^s$ of $n\geq
0$ temporal abstract
features $f_1^t,\ldots ,f_n^t$ and one spatial concrete feature $g^s$. 
An aspatial (concrete) feature chain is any finite composition
$f_1^{at}\ldots f_n^{at}g^{as}$ of $n\geq 0$ atemporal abstract features
$f_1^{at},\ldots ,f_n^{at}$ and one aspatial concrete feature
$g^{as}$.\footnote{Throughout the rest of the paper, a feature chain
$f_1\ldots f_kg$, either aspatial or spatial, is interpreted as within
the Description Logics Community ---i.e., as the composition
$f_1\circ\ldots\circ f_k\circ g$: we remind the reader that
$(f_1\circ\ldots\circ f_k\circ g)(x)=g(f_k(f_{k-1}(\ldots (f_2(f_1(x))))))$.}
The set of $\xdlplus$ concepts is the union of the the set of atemporal
concepts and the set of temporal concepts, which are the
smallest sets such that:
\begin{enumerate}
  \item\label{defxdlconceptsone} $\top$ and $\bot$ are
    atemporal concepts;
  \item\label{defxdlconceptstwo} $\top$ and $\bot$ are
    temporal concepts;\footnote{We could have used
    $\top ^{at}$ and $\top ^t$ for atemporal top and temporal top, respectively;
    and, similarly, $\bot ^{at}$ and $\bot ^t$ for atemporal bottom and
    temporal bottom, respectively.}
  \item\label{defxdlconceptsthree} an atemporal concept name is an atemporal
    concept;
  \item\label{defxdlconceptsfour} a temporal concept name is a temporal
    concept;
  \item\label{defxdlconceptsfive} if
    $C^{at}$ and $D^{at}$ are atemporal concepts;
    $C^t$ and $D^t$ are temporal concepts;
    $R^{at}$ is an atemporal role (in general, and an atemporal abstract feature in particular);
    $R^t$ is a temporal role (in general, and a temporal abstract feature in particular);
    $u_1^{as},\ldots ,u_n^{as}$ are aspatial feature chains;
    $u_1^t,u_2^t,u_3^t$ are spatial feature chains;
    $P^{as}$ is an aspatial $n$-ary predicate; and
    $P^s$ is a spatial predicate (binary if $x=\rcc8$, ternary if $x=\atra$),
    then:
    \begin{enumerate}
      \item $\neg C^{at}$,
            $C^{at}\sqcap D^{at}$,
            $C^{at}\sqcup D^{at}$,
            $\exists R^{at}.C^{at}$,
            $\forall R^{at}.C^{at}$ are atemporal concepts;
      \item $\exists (u_1^{as})\ldots (u_n^{as}).P^{as}$ is an atemporal concept;
      \item $\neg C^t$,
            $C^t\sqcap D^t$,
            $C^t\sqcup D^t$,
            $\exists R^t.C^t$,
            $\forall R^t.C^t$ are temporal concepts;
      \item $\exists (u_1^s)(u_2^s).P^s$, if $x$ binary,
            $\exists (u_1^s)(u_2^s)(u_3^s).P^s$, if $x$ ternary, are
            temporal concepts; and
      \item $\exists R^t.C^{at}$,
            $\forall R^t.C^{at}$ are temporal concepts.
    \end{enumerate}
\end{enumerate}
\end{definition}
$\alcd$ is the atemporal sublanguage of $\xdlplus$, and is generated
by Items \ref{defxdlconceptsone}, \ref{defxdlconceptsthree},
\ref{defxdlconceptsfive}(a) and \ref{defxdlconceptsfive}(b) of
Definition \ref{defxdlconcepts}. The spatio-temporalisation $\xdl$
we have already alluded to is the purely temporal part of $\xdlplus$, and is generated
by Items \ref{defxdlconceptstwo}, \ref{defxdlconceptsfour},
\ref{defxdlconceptsfive}(c) and \ref{defxdlconceptsfive}(d) of
Definition \ref{defxdlconcepts}.
We denote by $\mtalc$ the sublanguage of $\xdlplus$ given by rules
\ref{defxdlconceptstwo}, \ref{defxdlconceptsfour} and
\ref{defxdlconceptsfive}(c) in Definition \ref{defxdlconcepts},
which is the modal temporal logic component of $\xdlplus$. It is worth noting that $\mtalc$
does not consist of a mere temporalisation of $\alc$
\cite{Schmidt-Schauss91a}. Indeed, $\alc$ contains only general, not
necessarily functional roles,
whereas $\mtalc$ contains abstract features as well. As it will become
clear shortly, a mere temporalisation of $\alc$ (i.e., $\mtalc$
without abstract features) cannot capture the expressiveness of a
well-known modal temporal logic: Propositional Linear Temporal Logic
$\pltl$ \cite{Vardi86a}.
Given two integers $p\geq 0$ and $q\geq 0$, the sublanguage of
$\xdlplus$ (resp. $\mtalc$) whose concepts involve at most $p$
general, not necessarily functional temporal roles, and $q$ temporal abstract features
will be referred to as $\xdlpluspq$ (resp. $\mtalcpq$). We discuss shortly
the case $(p,q)=(0,1)$. We first
define weakly cyclic TBoxes.
\section{Weakly cyclic TBoxes}\label{sectsix}
An ($\xdlplus$ terminological) axiom is an expression of the form
$A\doteq C$, such that either (1) $A$ is an atemporal (defined) concept
name and $C$ an atemporal concept, or (2) $A$ is a temporal (defined)
concept name and $C$ a temporal concept. A TBox is a finite set of axioms,
with the condition that no concept name appears more than once as the left
hand side of an axiom.

Let $T$ be a TBox. $T$ contains two kinds of concept names: concept names appearing
as the left hand side of an axiom of $T$ are defined concepts; the others are
primitive concepts. A defined concept $A$ ``{\em directly uses}'' a defined
concept $B$ if and only if ($\iff$) $B$ appears in the right hand side of the axiom defining $A$. If
``{\em uses}'' is the transitive closure of ``{\em directly uses}'' then $T$
contains a cycle $\iff$ there is a defined concept $A$ that ``{\em uses}'' itself.
$T$ is cyclic if it contains a cycle; it is acyclic otherwise. $T$ is
weakly cyclic if it satisfies the following two conditions:
\begin{enumerate}
  \item Whenever $A$ uses $B$ and $B$ uses $A$, we have $B=A$ ---the only possibility for a defined concept to get
involved in a cycle is to appear in the right hand side of the axiom
defining it.
  \item All possible occurrences of a defined concept $B$ in the right
    hand side of the axiom defining $B$ itself, are within the scope of an
    existential or a universal quantifier; i.e., in subconcepts of 
    $C$ of the form $\exists R.D$ or $\forall R.D$, $C$ being the right 
    hand side of the axiom, $B\doteq C$, defining $B$.
\end{enumerate}
The TBox $T$ is temporally weakly cyclic and atemporally acyclic (or
$\twcatac$, for short) if it is weakly cyclic and, whenever
a defined concept $A$ uses itself, $A$ is a temporal defined concept.
Our intuition behind the use of $\twcatac$ TBoxes is to capture, on
the one hand, the expressiveness of $\alcd$ with acyclic TBoxes,
well-suited for the representation of static structured data and known to
be decidable, and, on the other hand, the expressiveness of $\xdl$ with
weakly cyclic TBoxes, which subsumes existing modal temporal logics while
remaining decidable -$\alcd$ with cyclic TBoxes is known to be undecidable.
As such, $\twcatac$ TBoxes are well-suited for the representation of change
in dynamic structured data.
We suppose that the temporal defined concepts of a TBox split into {\em eventuality}
defined concepts and {\em noneventuality} defined concepts.

In the rest of the paper, unless explicitly stated otherwise, we denote
concepts reducing to concept names by the letters $A$ and $B$,
possibly complex concepts by the letters $C$, $D$, $E$,
general (possibly functional) roles by the letter $R$,
abstract features by the letter $f$, concrete features by the letters $g$ and $h$,
feature chains by the letter $u$, predicates by the letter $P$. If distinguishing
between ``atemporal'' and ``temporal'' (resp. ``aspatial'' and ``spatial'') is
needed, we make use, as in Definition \ref{defxdlconcepts}, of the prefixes 'at'
and 't' (resp. 'as' and 's').
\section{Semantics of $\xdlplus$,
            with $x\in\{\rcc8 ,\atra\}$}\label{secteight}
$\xdlplus$ is equipped with a Tarski-style, possible worlds
semantics. $\xdlplus$ interpretations are spatio-temporal tree-like structures,
together with an interpretation
function associating with each temporal primitive concept $A$ the nodes of $t$
at which $A$ is true, and, additionally, associating with each
spatial concrete feature $g$ and each node $v$ of $t$, the value at $v$ (seen as
a time instant) of the spatial concrete object referred to by $g$. The
interpretation function also associates with each node of $t$ an $\alcd$
interpretation, which is a tree-like structure representing structured data
consisting of the situation (snapshot) of the World at the node (but excluding
the situation of the temporal primitive concepts and the relational spatial
situation, which are given by the temporal primitive concepts true at the node,
and the spatial concrete values associated with the spatial concrete features at
the node). Formally:
\begin{definition}[interpretation]
Let $x\in\{\rcc8 ,\atra\}$.
An interpretation ${\cal I}$ of $\xdlplus$ consists of a pair
${\cal I}=(t _{{\cal I}},.^{{\cal I}})$, where $t _{{\cal I}}$ is the domain of
${\cal I}$, consisting of a set of time points (or worlds, or states, or nodes), and
$.^{{\cal I}}$ is an interpretation function mapping each temporal primitive concept $A$ to a subset
$A^{{\cal I}}$ of $t_{{\cal I}}$, each temporal role $R$ to a subset $R^{{\cal I}}$ of
$t _{{\cal I}}\times t _{{\cal I}}$, so that $R^{{\cal I}}$
is functional if $R$ is an abstract feature, and each
spatial concrete feature $g$ to a total function $g^{{\cal I}}$:
\begin{footnotesize}
\begin{enumerate}
  \item from $t _{{\cal I}}$ onto the set $\rtopspace$ of regions of a topological space $\topspace$, if $x=\rcc8$;
	 and
  \item from $t _{{\cal I}}$ onto the set $\deuxdo$ of orientations of the 2-dimensional
    space, if $x=\atra$.
\end{enumerate}
\end{footnotesize}
Each temporal role $R$ should be so that the reflexive-transitive closure
$(R^{{\cal I}})^*$ of $R^{{\cal I}}$ is serial and antisymmetric, making
interpretation ${{\cal I}}$ a branching tree-like temporal structure. The
interpretation function $.^{{\cal I}}$ also associates with each time
point $v$ in $t_{{\cal I}}$ an $\alcd$ interpretation
$.^{{\cal I}}(v)=(\Delta _{{\cal I},v},.^{{\cal I},v})$, where
$\Delta _{{\cal I},v}$ is a set consisting of the (abstract) domain of
$.^{{\cal I}}(v)$ and $.^{{\cal I},v}$ is an interpretation function
mapping each atemporal concept name $C$ (either defined or primitive)
to a subset $C^{{\cal I},v}$ of $\Delta _{{\cal I},v}$, each atemporal
role $R$ to a subset $R^{{\cal I},v}$ of $\Delta
_{{\cal I},v}\times\Delta _{{\cal I},v}$, each aspatial
concrete feature $g$ to a partial function $g^{{\cal I},v}$ from
$\Delta _{{\cal I},v}$ onto the set $\Delta _{{\cal D}}$ of concrete objects of the aspatial concrete domain ${\cal D}$.
The interpretation function $.^{{\cal I},v}$ is extended to arbitrary atemporal concepts as follows:
\begin{footnotesize}
\begin{eqnarray}
(\top )^{{\cal I},v}&:=&\Delta _{{\cal I},v}  \nonumber\\
(\bot )^{{\cal I},v}&:=&\emptyset  \nonumber\\
(\neg C)^{{\cal I},v}&:=&\Delta _{{\cal I},v}\setminus C^{{\cal I},v}  \nonumber\\
(C\sqcap D)^{{\cal I},v}&:=&C^{{\cal I},v}\cap D^{{\cal I},v}  \nonumber\\
(C\sqcup D)^{{\cal I},v}&:=&C^{{\cal I},v}\cup D^{{\cal I},v}  \nonumber\\
(\exists R.C)^{{\cal I},v}&:=&\{a\in\Delta _{{\cal I},v}|  \nonumber\\
                          &  &\exists
    b\in\Delta _{{\cal I},v}:(a,b)\in R^{{\cal I},v}\wedge b\in C^{{\cal I},v}\}  \nonumber\\
(\forall R.C)^{{\cal I},v}&:=&\{a\in\Delta _{{\cal I},v}|  \nonumber\\
                          &  &\forall b:(a,b)\in R^{{\cal I},v}\rightarrow b\in C^{{\cal I},v}\}  \nonumber\\
(\exists (u_1)\ldots (u_n).P)^{{\cal I},v}&:=&\{a\in\Delta
    _{{\cal I},v}|  \nonumber\\
                                          &  &\exists o_1,\ldots o_n\in\Delta _{{\cal D}} :
                    u_1^{{\cal I},v}(a)=o_1,\ldots ,  \nonumber\\
                              &&
                    u_n^{{\cal I},v}(a)=o_n,
                    P(o_1,\ldots ,o_n)\}  \nonumber
\end{eqnarray}
\end{footnotesize} 
where, given an aspatial feature chain
$u=f_1\ldots f_ng$, $u^{{\cal I},v}(a)$ stands for
the value $g^{{\cal I},v}(b)$, where $b$ is the
$f_1^{{\cal I},v}\ldots f_n^{{\cal I},v}$-successor of $a$ in the $\alcd$
interpretation $.^{{\cal I}}(v)$.
\end{definition}
\begin{definition}[satisfiability $\wrt$ a TBox]
Let $x\in\{\rcc8 ,\atra\}$ be a spatial RA, $C$ an $\xdlplus$ concept, ${\cal T}$ an $\xdlplus$
$\twcatac$ TBox, and ${\cal I}=(t_{{\cal I}},.^{{\cal I}})$ an
$\xdlplus$ interpretation. The satisfiability, by a node $s$ of $t_{{\cal
    I}}$, of $C$ $\wrt$ to ${\cal T}$, denoted ${\cal I},s\models\langle C,{\cal T}\rangle$, is defined inductively as follows (Item 1. below deals with the case of an atemporal concept, the remaining 11 with a temporal concept):
\begin{enumerate}
  \item ${\cal I},s\models\langle C,{\cal T}\rangle$ $\iff$ $C^{{\cal I},v}\not =\emptyset$,
	for all atemporal concepts $C$
  \item ${\cal I},s\models\langle\top ,{\cal T}\rangle$
  \item ${\cal I},s\not\models\langle\bot ,{\cal T}\rangle$
  \item ${\cal I},s\models\langle A,{\cal T}\rangle$ $\iff$
    $s\in A^{{\cal I}}$, for all primitive concepts $A$
  \item ${\cal I},s\models\langle B,{\cal T}\rangle$ $\iff$
    ${\cal I},s\models\langle C,{\cal T}\rangle$, for all defined
    concepts $B$ given by the axiom $B\doteq C$ of ${\cal T}$
  \item ${\cal I},s\models\langle\neg C,{\cal T}\rangle$ $\iff$
    ${\cal I},s\not\models\langle C,{\cal T}\rangle$
  \item ${\cal I},s\models\langle C\sqcap D,{\cal T}\rangle$ $\iff$
    ${\cal I},s\models\langle C,{\cal T}\rangle$ and ${\cal
      I},s\models\langle D,{\cal T}\rangle$
  \item ${\cal I},s\models\langle C\sqcup D,{\cal T}\rangle$ $\iff$
    ${\cal I},s\models\langle C,{\cal T}\rangle$ or ${\cal
      I},s\models\langle D,{\cal T}\rangle$
  \item ${\cal I},s\models\langle\exists R.C,{\cal T}\rangle$ $\iff$
    ${\cal I},s'\models\langle C,{\cal T}\rangle$, for some $s'$ such
    that $(s,s')\in R^{{\cal I}}$
  \item ${\cal I},s\models\langle\forall R.C,{\cal T}\rangle$ $\iff$
    ${\cal I},s'\models\langle C,{\cal T}\rangle$, for all $s'$ such
    that $(s,s')\in R^{{\cal I}}$
  \item ${\cal I},s\models\langle\exists (u_1)(u_2).P,{\cal T}\rangle$ $\iff$
    $P(u_1^{{\cal I}}(s),u_2^{{\cal I}}(s))$
  \item ${\cal I},s\models\langle\exists (u_1)(u_2)(u_3).P,{\cal T}\rangle$ $\iff$
    $P(u_1^{{\cal I}}(s),u_2^{{\cal I}}(s),u_3^{{\cal I}}(s))$
\end{enumerate}
A concept $C$ is satisfiable $\wrt$ a TBox ${\cal T}$ 
$\iff$ ${\cal I},s\models\langle C,{\cal T}\rangle$, for some $\xdlplus$
interpretation ${\cal I}$, and some state
$s\in t_{{\cal I}}$, in which case the pair $({\cal I},s)$ is a model
of $C$ $\wrt$ ${\cal T}$; $C$ is insatisfiable (has no models) $\wrt$ ${\cal T}$,
otherwise. $C$ is valid $\wrt$ ${\cal T}$ $\iff$ the negation, $\neg
C$, of $C$ is insatisfiable $\wrt$ ${\cal T}$. The satisfiability
problem and the subsumption problem are defined as follows:
\begin{enumerate}
  \item[] {\em The satisfiability problem:} given a concept $C$ and a
	TBox ${\cal T}$, is $C$ satisfiable $\wrt$ ${\cal T}$?
  \item[] {\em The subsumption problem:} given two concepts $C$ and $D$ and a TBox ${\cal T}$,
      does $C$ subsume $D$ $\wrt$ ${\cal T}$ (notation:
      $D\sqsubseteq _{{\cal T}}C$)? in other words, are all models of $D$ $\wrt$
        ${\cal T}$ also models of $C$ $\wrt$ ${\cal T}$?
\end{enumerate}
\end{definition}
The satisfiability problem and the subsumption problem are related to
each other, as follows: $D\sqsubseteq _{{\cal T}}C$ $\iff$
$D\sqcap\neg C$ is insatisfiable $\wrt$ ${\cal T}$.
\section{Associating a weak alternating automaton with the satisfiability of an $\xdl$ concept $\wrt$ a
weakly cyclic TBox: an overview}\label{sectnine}
It should be clear that, given decidability of the satisfiability of an $\alcd$ concept $\wrt$ an acyclic TBox,
in order to show decidability of the satisfiability of an $\xdlplus$ concept $\wrt$ a $\twcatac$ TBox, it is
sufficient to show decidability of an $\xdl$ concept $\wrt$ a weakly cyclic TBox. The following is an overview
of a proof of such a decidability. Given an $\xdl$ concept
$C$ and an $\xdl$ weakly cyclic TBox ${\cal T}$, the problem we are interested in
is, the satisfiability of $C$ with respect to ${\cal T}$. The axioms in ${\cal T}$ are of the form $B\doteq E$,
where $B$ is a defined concept name, and $E$ an $\xdl$ concept. Using $C$, we introduce a new defined
concept name, $B_{init}$, given by the axiom $B_{init}\doteq C$. We denote by ${\cal T}'$ the TBox consisting
of ${\cal T}$ augmented with the new axiom: ${\cal T}'={\cal T}\cup\{B_{init}\doteq C\}$. The alternating automaton we
associate with the satisfiability of $C$ $\wrt$ the TBox ${\cal T}$, so that satisfiability holds $\iff$ the
language accepted by the automaton is not empty, is now almost entirely given by the TBox ${\cal T}'$: the
defined concept names represent the states of the automaton, $B_{init}$ being the initial state; the
transition function is given by the axioms themselves. However, some modification of the axioms is needed.

Given an $\xdl$ axiom $B\doteq E$ in ${\cal T}'$, the method we propose decomposes $E$ into some kind of
Disjunctive Normal Form, $\dnf2 (E)$, which is free of occurrences of
the form $\forall R.E'$. Intuitively, the concept $E$ is satisfiable by the state consisting
of the defined concept name $B$, $\iff$ there exists an element $S$ of $\dnf2 (E)$ that is satisfiable by
$B$. An element $S$ of $\dnf2 (E)$ is a conjunction written as a set, of the form
$S_{prop}\cup S_{csp}\cup S_{\exists}$, where:
\begin{enumerate}
  \item $S_{prop}$ is a set of primitive concepts and negated primitive concepts ---it is worth
noting here that, while the defined concepts (those concept names appearing as the left hand side of an axiom) define the
states of our automaton, the primitive concepts (the other concept names) correspond to atomic propositions in, e.g.,
classical propositional calculus;
  \item $S_{csp}$ is a set of concepts of the form $\exists (u_1)\cdots (u_n).P$,
    where $u_1,\ldots ,u_n$ are feature chains and $P$ a relation (predicate) of an $n$-ary
    spatial RA; and
  \item $S_{\exists}$ is a set of concepts of the form $\exists
    R.E_1$, where $R$ is a role and $E_1$ is a concept.
\end{enumerate}
The procedure ends
with a TBox ${\cal T}'$ of which all axioms are so written. Once ${\cal T}'$ has been so written, we denote:
\begin{enumerate}
  \item by $af({\cal T}')$, the set of abstract features appearing in ${\cal T}'$; and
  \item by $rrc({\cal T}')$, the set of concepts appearing in ${\cal T}'$, of the form $\exists R.E$, with $R$ being a
    general, not necessarily functional role, and $E$ a concept.
\end{enumerate}
The alternating automaton to be associated with ${\cal T}'$, will operate on (Kripke) structures which are
infinite $m+p$-ary trees, with $m=|af({\cal T}')|$ and $p=|rrc({\cal T}')|$. Such a structure, say $t$, is associated
with a truth-value assignment function $\pi$, assigning to each node, the set of those primitive concepts
appearing in ${\cal T}'$ that are true at the node. With $t$ are also associated the concrete features
appearing in ${\cal T}'$: such a concrete feature, $g$, is mapped at each node of $t$, to a (concrete) object
of the spatial domain in consideration (e.g., a region of a topological space if the concrete domain is
generated by $\rcc8$).

The feature chains are of the form $f_1\ldots f_kg$, with $k\geq 0$, where the
$f_i$'s are abstract features (also known, as alluded to before, as functional roles: functions from
the abstract domain onto the abstract domain), whereas $g$ is a concrete feature (a function from the
abstract domain onto the set of objects of the concrete domain). The sets $S$ are used to label the
nodes of the search space. Informally, a run of the \mbox{tableaux-like} search space is a
\mbox{disjunction-free} subspace, obtained by selecting at each node, labelled, say, with $S$, one
element of $\dnf2 (S)$.

Let $\sigma$ be a run, $s_0$ a node of $\sigma$, and $S$ the label of $s_0$, and suppose that $S_{csp}$
contains $\exists (u_1)(u_2).P$ (we assume, without loss of generality, a concrete domain generated by
a binary spatial RA, such as $\rcc8$ \cite{Randell92a}), with
$u_1=f_1\ldots f_kg_1$ and $u_2=f _1'\ldots f _m'g_2$.
The concept $\exists (u_1)(u_2).P$ gives birth to new nodes of the run,
$s_1=f_1(s_0),s_2=f_2(s_1),\ldots ,s_k=f_k(s_{k-1}),
 s_{k+1}=f _1'(s_0),s_{k+2}=f _2'(s_{k+1}),\ldots ,s_{k+m}=f _m'(s_{k+m-1})$; to new variables of
what could be called the (global) CSP, $\csp (\sigma )$, of $\sigma$; and to a new constraint of
$\csp (\sigma )$. The new variables are $\langle s_k,g_1\rangle$ and
$\langle s_{k+m},g_2\rangle$, which denote the values of the
concrete features $g_1$ and $g_2$ at nodes $s_k$ and $s_{k+m}$, respectively. The new constraint is
$P(\langle s_k,g_1\rangle ,\langle s_{k+m},g_2\rangle)$.
The set of all such variables together with the set of all such constraints, generated by node $s_0$,
give the $\csp$ $\csp _{\sigma}(s_0)$ of $\sigma$ at $s_0$; and the union of all
$\csp$s $\csp _s({\sigma})$, over the nodes $s$ of $\sigma$, gives $\csp (\sigma )$. The
feature chains make it possible to refer to the values of the
different concrete features at the different nodes of a run, and restrict
these values using spatial predicates.

The pruning process during the tableaux method will now work as follows. The search will make use
of a data structure $\queue$, which will be handled in very much the same fashion as such a data
structure is handled in local consistency algorithms, such as arc- or path-consistency
in standard $\csp$s. The data structure is initially empty. Then
whenever a new node $s$ is added to the search space, the global $\csp$ of the run being
constructed is updated, by augmenting it with (the variables and) the constraints generated, as
described above, by $s$. Once the $\csp$ has been updated, so that it includes the local $\csp$ at
the current node, the local consisteny pruning is applied by propagating the constraints in
$\queue$.
Once a run has been fully
constructed, and only then, its global $\csp$ is solved. In the case of a concrete
domain generated by a binary, $\rcc8$-like RA, the filtering is achieved with a path-consisteny
algorithm \cite{Allen83b}, and the solving of the global $\csp$, after a run has been fully
constructed, with a solution search algorithm such as the one in \cite{Ladkin92a}. In the
case of a concrete domain generated by a ternary spatial RA, the filtering and the solving
processes are achieved with a strong 4-consistency and a search algorithms such as the ones in \cite{Isli00b}.
\section{Summary}
We have provided a rich spatio-temporal framework combining a
spatio-temporalisation of the well-know $\alcd$ family of
description logics with a concrete domain \cite{Baader91a},
with $\alcd$ itself. The famework is well-suited for the
representation of change in dynamic structured data, in dynamic
spatial scenes, and in dynamic propositional knowledge. Contrary
to most existing approaches of combining modal or description
logics to get spatio-temporal languages (see, e.g.,
\cite{Balbiani02a,Bennett02a,Bennett02b,Wolter00a}), ours leads to a decidable
language. This advantage of being expressively rich while remaining decidable
is the fruit of the way the combination is done, which is complex enough to
make the resulting framework rich, but keeps a separation between the
(decidable) combined languages large enough to bring decidability of the
resulting language into decidability of the combined ones.
\bibliography{}
\begin{center}
THE NOTIFICATION LETTER\\
(as received on 3 May 2004)
\end{center}
Dear Amar Isli:

We regret to inform you that your submission 

  C0686
  Augmenting ALC(D) (atemporal) roles and (aspatial) concrete domain
  with temporal roles and a spatial concrete domain -first results
  Amar Isli
 
cannot be accepted for inclusion in the ECAI 2004's programme. Due to 
the large number of submitted papers, we are aware that also otherwise 
worthwhile papers had to be excluded. You may then consider submitting 
your contribution to one of the ECAI's workshops, which are still open 
for submission.

In this letter you will find enclosed the referees' comments on your 
paper.

We would very much appreciate your participation in the meeting and 
especially in the discussions. 

Please have a look at the ECAI 2004 website for registration details 
and up-to-date information on workshops and tutorials:

    http://www.dsic.upv.es/ecai2004/

The schedule of the conference sessions will be available in May 2004. 

I thanks you again for submitting to ECAI 2004 and look forward to 
meeting you in Valencia.

Best regards
      
Programme Committee Chair
\begin{center}
REVIEW ONE
\end{center}
----- ECAI 2004 REVIEW SHEET FOR AUTHORS -----

PAPER NR: C0686 

TITLE: Augmenting ALC(D) (atemporal) roles and (aspatial) concrete 
domain
       with temporal roles and a spatial concrete domain -first
       results

1) SUMMARY (please provide brief answers)

- What is/are the main contribution(s) of the paper?

No substantial results and contribution.

2) TYPE OF THE PAPER

The paper reports on:

  [X] Preliminary research

  [ ] Mature research, but work still in progress

  [ ] Completed research

The emphasis of the paper is on:

  [ ] Applications

  [X] Methodology

3) GENERAL RATINGS

Please rate the 6 following criteria by, each time, using only 
one of the five following words: BAD, WEAK, FAIR, GOOD, EXCELLENT

3a) Relevance to ECAI: FAIR

3b) Originality: WEAK

3c) Significance, Usefulness: BAD

3d) Technical soundness: FAIR

3e) References: BAD

3f) Presentation: WEAK

4) QUALITY OF RESEARCH

4a) Is the research technically sound?

    [ ] Yes   [X] Somewhat   [ ] No

4b) Are technical limitations/difficulties adequately discussed?

    [ ] Yes   [ ] Somewhat   [X] No

4c) Is the approach adequately evaluated?

    [ ] Yes   [ ] Somewhat   [X] No

FOR PAPERS FOCUSING ON APPLICATIONS:

4d) Is the application domain adequately described?

    [ ] Yes   [ ] Somewhat   [ ] No

4e) Is the choice of a particular methodology discussed?

    [ ] Yes   [ ] Somewhat   [ ] No

FOR PAPERS DESCRIBING A METHODOLOGY:

4f) Is the methodology adequately described?

    [ ] Yes   [ ] Somewhat   [X] No

4g) Is the application range of the methodology adequately described, 
    e.g. through clear examples of its usage?

    [ ] Yes   [ ] Somewhat   [X] No

Comments:

See below.

5) PRESENTATION

5a) Are the title and abstract appropriate? 

    [ ] Yes   [X] Somewhat   [ ] No

5b) Is the paper well-organized? [ ] Yes   [ ] Somewhat   [X] No

5c) Is the paper easy to read and understand? 

    [ ] Yes   [ ] Somewhat   [X] No

5d) Are figures/tables/illustrations sufficient? 

    [ ] Yes   [ ] Somewhat   [X] No

5e) The English is  [X] very good   [ ] acceptable   [ ] dreadful

5f) Is the paper free of typographical/grammatical errors? 

    [X] Yes   [ ] Somewhat   [ ] No

5g) Is the references section complete?

    [ ] Yes   [ ] Somewhat   [X] No

Comments:

See below.

6) TECHNICAL ASPECTS TO BE DISCUSSED (detailed comments)

- Suggested / required modifications:

General comments.

At the beginning of the paper, many elements are introduced at a very
abstract level, with a very few explanations (and nothing is said
about the supposed knowledge of the reader with respect to the
subject, and the paper is far from being self-explanatory).
Moreover, the reader does not understand on which basis he must rely
for reading the paper, and what is the actual objective of the paper.

There is a real problem with this paper: the author(s) propose a
number of definitions (not always easy to read and to understand)
without giving any concrete example.
There is no discussion on the framework that they have introduced,
and no comparison with related papers on space and time in description
logics (such as V. Haarslev, C. Lutz and R. Moeller, A Description
Logic with Concrete Domains and a Role-forming Predicate Operator,
Journal of Logic and Computation, 9(3):351-384, 1999).
In these conditions, it is very difficult to have a good idea of the
objectives of the paper, and the utility of the paper: this is perhaps
a nice theoretical work, but what can we do with this framework?
How can we really take into account in practical situations time and
space for solving real-world problems?
For terminating, what is the actual objective of the author(s) writing
this paper?
The reader is not convinced, and the paper is not acceptable under its
present form.

- Other comments:

\begin{center}
REVIEW TWO
\end{center}
----- ECAI 2004 REVIEW SHEET FOR AUTHORS -----

PAPER NR: C0686 

TITLE: Augmenting ALC(D) (atemporal) roles and (aspatial) concrete 
domain with temporal roles and a spatial concrete domain - first results

1) SUMMARY (please provide brief answers)

- What is/are the main contribution(s) of the paper?

The paper describes the spatio-temporalisation of the ALC(D) family of 
description logics.

2) TYPE OF THE PAPER

The paper reports on:

  [X] Preliminary research

  [ ] Mature research, but work still in progress

  [ ] Completed research

The emphasis of the paper is on:

  [ ] Applications

  [X] Methodology

3) GENERAL RATINGS

Please rate the 6 following criteria by, each time, using only 
one of the five following words: BAD, WEAK, FAIR, GOOD, EXCELLENT

3a) Relevance to ECAI: FAIR

3b) Originality:  

3c) Significance, Usefulness: FAIR

3d) Technical soundness: FAIR

3e) References: WEAK

3f) Presentation: BAD

4) QUALITY OF RESEARCH

4a) Is the research technically sound?

    [ ] Yes   [X] Somewhat   [ ] No

4b) Are technical limitations/difficulties adequately discussed?

    [ ] Yes   [ ] Somewhat   [X] No

4c) Is the approach adequately evaluated?

    [ ] Yes   [ ] Somewhat   [X] No

FOR PAPERS FOCUSING ON APPLICATIONS:

4d) Is the application domain adequately described?

    [ ] Yes   [ ] Somewhat   [ ] No

4e) Is the choice of a particular methodology discussed?

    [ ] Yes   [ ] Somewhat   [ ] No

FOR PAPERS DESCRIBING A METHODOLOGY:

4f) Is the methodology adequately described?

    [ ] Yes   [X] Somewhat   [ ] No

4g) Is the application range of the methodology adequately described, 
    e.g. through clear examples of its usage?

    [ ] Yes   [ ] Somewhat   [X] No

Comments:

The quality of presentation of the paper is not sufficient to make a 
reliable judgment regarding the general quality of the research, hence 
the largely neutral ratings of this section.

5) PRESENTATION

5a) Are the title and abstract appropriate? 

    [ ] Yes   [ ] Somewhat   [X] No

5b) Is the paper well-organized? [ ] Yes   [ ] Somewhat   [X] No

5c) Is the paper easy to read and understand? 

    [ ] Yes   [ ] Somewhat   [X] No

5d) Are figures/tables/illustrations sufficient? 

    [ ] Yes   [ ] Somewhat   [X] No

5e) The English is  [ ] very good   [X] acceptable   [ ] dreadful

5f) Is the paper free of typographical/grammatical errors? 

    [ ] Yes   [ ] Somewhat   [X] No

5g) Is the references section complete?

    [ ] Yes   [X] Somewhat   [ ] No

Comments:

The presentation of this work lets it down completely. It is below the 
standard necessary for a general international audience of AI 
researchers, and this virtually debars it from the possibility of a measured 
technical evaluation. The paper tries to cram far too much technical 
detail into too little space, at the expense of any high-level, informal or 
intuitive description of the work, or any detailed indication of its 
applicability. There is not a single example to aid comprehension or 
readability.

6) TECHNICAL ASPECTS TO BE DISCUSSED (detailed comments)

- Suggested / required modifications:

To be acceptable for publication within the given page limitation, this 
work needs to be described (at least partly) at a more informal and 
intuitive level, and with the aid of examples.

- Other comments:

\begin{center}
REVIEW THREE
\end{center}
----- ECAI 2004 REVIEW SHEET FOR AUTHORS -----

PAPER NR: C0686 

TITLE: Augmenting ALC(D) (atemporal) roles and (aspatial)....

1) SUMMARY (please provide brief answers)

- What is/are the main contribution(s) of the paper?

2) TYPE OF THE PAPER

The paper reports on:

  [ ] Preliminary research

  [ ] Mature research, but work still in progress

  [ ] Completed research

The emphasis of the paper is on: X

  [ ] Applications

  [ ] Methodology

3) GENERAL RATINGS

Please rate the 6 following criteria by, each time, using only 
one of the five following words: BAD, WEAK, FAIR, GOOD, EXCELLENT

3a) Relevance to ECAI:  

3b) Originality:  

3c) Significance, Usefulness:  

3d) Technical soundness:  

3e) References:  

3f) Presentation:

4) QUALITY OF RESEARCH

4a) Is the research technically sound?

    [ ] Yes   [ ] Somewhat   [ ] No

4b) Are technical limitations/difficulties adequately discussed?

    [ ] Yes   [ ] Somewhat   [ ] No

4c) Is the approach adequately evaluated?

    [ ] Yes   [ ] Somewhat   [ ] No

FOR PAPERS FOCUSING ON APPLICATIONS:

4d) Is the application domain adequately described?

    [ ] Yes   [ ] Somewhat   [ ] No

4e) Is the choice of a particular methodology discussed?

    [ ] Yes   [ ] Somewhat   [ ] No

FOR PAPERS DESCRIBING A METHODOLOGY:

4f) Is the methodology adequately described?

    [ ] Yes   [ ] Somewhat   [ ] No

4g) Is the application range of the methodology adequately described, 
    e.g. through clear examples of its usage?

    [ ] Yes   [ ] Somewhat   [ ] No

Comments:

5) PRESENTATION

5a) Are the title and abstract appropriate? 

    [ ] Yes   [ ] Somewhat   [ ] No

5b) Is the paper well-organized? [ ] Yes   [ ] Somewhat   [ ] No

5c) Is the paper easy to read and understand? 

    [ ] Yes   [ ] Somewhat   [ ] No

5d) Are figures/tables/illustrations sufficient? 

    [ ] Yes   [ ] Somewhat   [ ] No

5e) The English is  [ ] very good   [ ] acceptable   [ ] dreadful

5f) Is the paper free of typographical/grammatical errors? 

    [ ] Yes   [ ] Somewhat   [ ] No

5g) Is the references section complete?

    [ ] Yes   [ ] Somewhat   [ ] No

Comments:

6) TECHNICAL ASPECTS TO BE DISCUSSED (detailed comments)

- Suggested / required modifications:

- Other comments:
\end{document}